\documentclass[letterpaper, 10 pt, conference]{ieeeconf}  
\usepackage{FG2026}
\FGfinalcopy

\usepackage{amsmath}
\usepackage{booktabs}
\usepackage{graphicx}
\usepackage[table]{xcolor}

\IEEEoverridecommandlockouts                              
\def\FGPaperID{340} 

\title{\LARGE \bf
Diffusion for De-Occlusion: Accessory-Aware Diffusion Inpainting for Robust Ear Biometric Recognition
}

\author{%
  \parbox{\textwidth}{\centering
    {\large Deeksha Arun$^{1}$, Kevin W. Bowyer$^{1}$, Patrick Flynn$^{1}$}\\
    {\normalsize
      $^{1}$Department of Computer Science and Engineering, University of Notre Dame, Notre Dame, IN 46556
    }\\
  }
}

\begin{document}

\ifFGfinal
\thispagestyle{empty}
\pagestyle{empty}
\else
\author{Anonymous FG2026 submission\\ Paper ID \FGPaperID \\}
\pagestyle{plain}
\fi
\maketitle

\begin{abstract}

Ear occlusions 
(arising from the presence of ear accessories such as earrings and earphones)
can negatively impact performance in ear-based biometric recognition systems, especially in unconstrained imaging circumstances. In this study, we assess the effectiveness of a diffusion-based ear inpainting technique as a pre-processing aid to mitigate the issues of ear accessory occlusions in transformer-based ear recognition systems. Given an input ear image and an automatically derived accessory mask, the inpainting model reconstructs clean and anatomically plausible ear regions by synthesizing missing pixels while preserving local geometric coherence along key ear structures, including the helix, antihelix, concha, and lobule. We evaluate the effectiveness of this pre-processing aid in transformer-based recognition systems for several vision transformer models and different patch sizes for a range of benchmark datasets. Experiments show that diffusion-based inpainting can be a useful pre-processing aid to alleviate ear accessory occlusions to improve overall recognition performance.

\end{abstract}

\begin{figure*}[t]
\centering
  \includegraphics[width=0.9\linewidth,clip=]{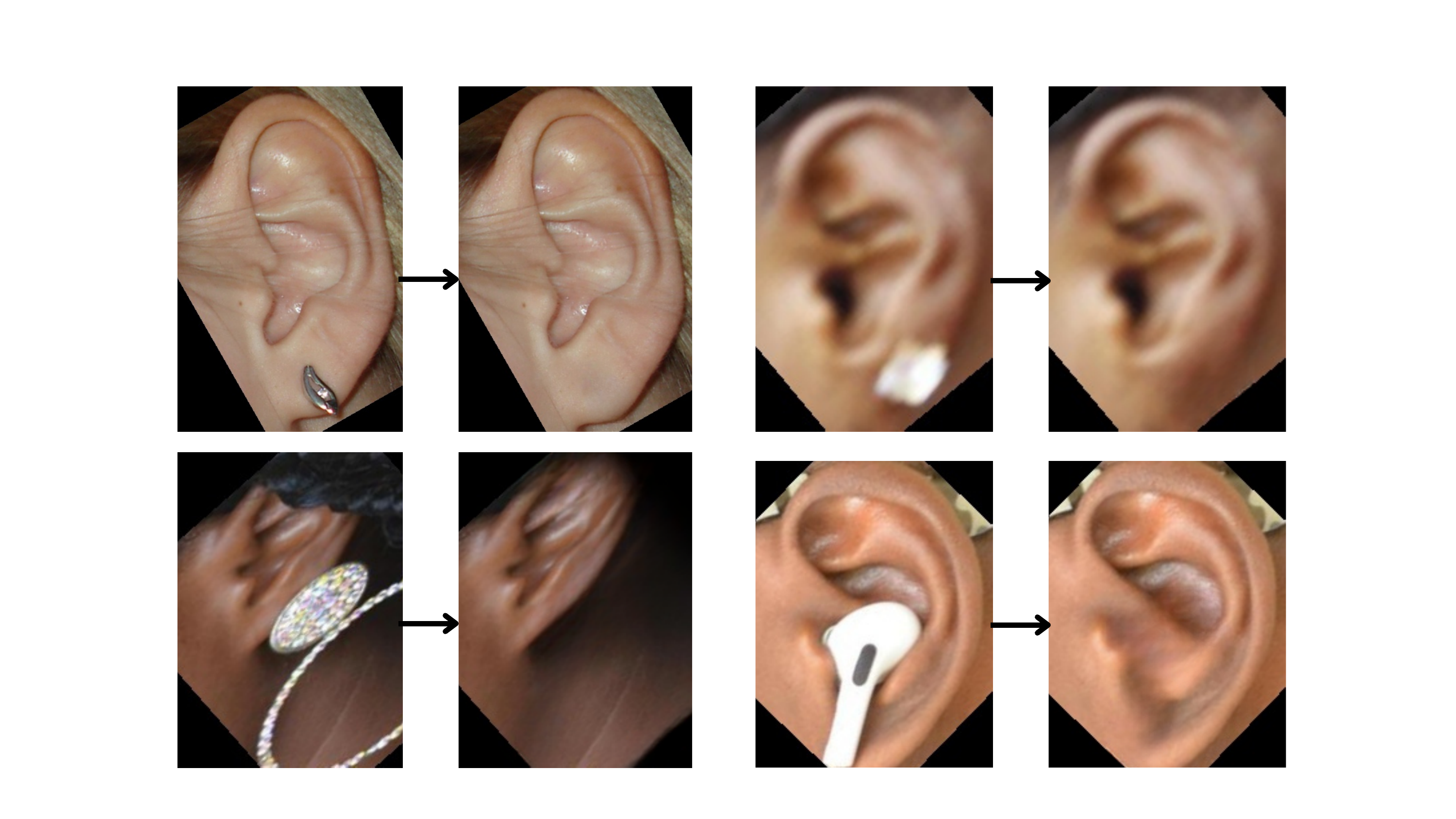}
  \caption{\textbf{Four representative ear images (original vs. inpainted) produced by the proposed Ear-accessory Inpainting pipeline}: Left indicate original aligned and cropped images (with accessories) and right indicates aligned and cropped + inpainted images (without accessories).}
  \label{fig:ear_images}
\end{figure*}
\section{INTRODUCTION}

Ear biometrics has attracted growing attention as a practical modality for unconstrained recognition due to the ear's rich structural cues and its relative permanence~\cite{sforza2009age, yoga2017assessment}. Compared with faces, ears are less affected by expression and makeup and can remain discriminative even under moderate pose variation, making them appealing for applications ranging from access control to forensic analysis~\cite{benzaoui2023comprehensive}. However, real-world deployments frequently encounter partial occlusions that disrupt the visible anatomy of the ear and undermine the stability of learned representations~\cite{emersic2017covariate}. Among these, ear accessories, particularly earrings and earphones, are increasingly common and pose a distinct challenge: they not only cover key regions but also introduce strong specular highlights and non-biological textures that bias feature extraction and matching. Accessories can serve as spurious identity cues since a model may latch onto them when a subject wears them during enrollment. This can cause missed matches when the accessory is absent or different, and false matches when another subject wears a similar accessory, enabling presentation attacks~\cite{emervsivc2018towards}.

Accessory-induced occlusions differ from typical nuisances such as hair or scarves in two important ways. First, earrings and earphones often overlap identity-relevant structures near the lobule, concha, and tragus, where curvature and shadow patterns contribute substantially to recognition. Second, these items vary widely in shape, material, and placement, producing high-frequency patterns and sharp boundaries that can dominate convolutional features and attention maps. As a result, recognition systems trained primarily on unobstructed ears often suffer significant accuracy drops when accessories are present, while naive strategies such as masking the occluded area discard information and can reduce discriminability even further.

A natural remedy is to restore the occluded ear region prior to recognition. Classical inpainting methods and early deep generative approaches~\cite{emervsivc2018towards} can fill missing pixels, but they often struggle to reconstruct fine anatomical details and may produce blurred or inconsistent structures, especially when the occlusion spans curved boundaries. These limitations motivate restoration methods that can generate sharp, structure-consistent reconstructions while minimizing identity drift.

Diffusion models~\cite{croitoru2023diffusion} have recently emerged as a strong class of generative models for image synthesis and editing, exhibiting notable robustness in producing high-fidelity, semantically coherent content. Their iterative denoising process allows them to refine local details while maintaining global consistency, making them particularly suited for inpainting tasks that require realistic completion under complex occlusions. In the context of ear biometrics, diffusion-based inpainting offers the opportunity to remove accessories and reconstruct plausible ear anatomy, but it must be adapted to satisfy domain-specific constraints: reconstructions should align with expected ear geometry, preserve fine-grained texture, and avoid introducing artifacts that could alter identity-relevant cues.

In this work, we investigate diffusion-based inpainting as a pre-processing strategy to mitigate ear accessory occlusions. Given an ear image with earrings or earphones, we first estimate the accessory region and then perform diffusion-based inpainting to synthesize the underlying ear content. We assess the approach through its impact on downstream recognition using Vision Transformers, since biometric utility is the central objective. Qualitative examples of the original and corresponding inpainted ear images produced by our pipeline are shown in Fig.~\ref{fig:ear_images}.

Our key contributions are as follows:
\begin{enumerate}
\item \textit{The first ear accessory-aware diffusion-based inpainting pipeline} that reconstructs occluded helix/antihelix/concha/lobule regions to produce accessory-free ear inputs.
\item \textit{The first fully automated mask generation pipeline} combining YOLOv10~\cite{wang2024yolov10}, Grounding DINO~\cite{liu2023grounding}, and SAM2~\cite{ravi2024sam2} to obtain robust ear accessory segmentations.
\item A \textit{multi-backbone, multi-tokenization evaluation} on four benchmarks (AWE \cite{emervsivc2017ear}, OPIB \cite{Adebayo2023}, WPUT \cite{frejlichowski2010west}, and EarVN1.0 \cite{hoang2019earvn1}) showing when restoration improves verification and analyzing failure cases due to masking errors and padding artifacts.

\end{enumerate}

Collectively, these results suggest that diffusion-driven restoration can serve as an effective front-end for robust ear recognition in real-world scenarios where accessories are prevalent.


\section{Related Work}

\subsection{Ear Biometrics and Occlusion Robustness}
Ear-based biometric recognition has been extensively studied due to the ear’s stable anatomical structure, rich geometric cues, and suitability for acquisition under profile views~\cite{frejlichowski2010west}. Early approaches relied on handcrafted features such as contour descriptors, local binary patterns, and geometric measurements of auricular components~\cite{dodge2018unconstrained, alshazly2019handcrafted, moreno1999use, choras2010ear}. With the advent of deep learning, convolutional neural networks and, more recently, vision transformers have substantially improved ear recognition accuracy by learning discriminative representations directly from data~\cite{arun2025,ying2018human, abdellatef2020fusion}.

Despite these advances, occlusion remains a major challenge for ear biometrics in unconstrained environments. Several studies have shown that occlusions caused by hair, scarves, and accessories can significantly degrade recognition performance~\cite{emervsivc2017ear, hoang2019earvn1}. Prior work on accessories-aware ear recognition analyzes how ear accessories can degrade recognition performance and be exploited for accessory-based presentation attacks, and it evaluates simple mitigation strategies such as masking (fixed/adaptive color) and CNN-based inpainting to remove accessory regions~\cite{emervsivc2018towards}. However, the study is conducted under oracle assumptions with ground-truth accessory masks, relies largely on synthetically superimposed earrings rather than diverse real-world accessories.

Accessory-induced occlusions, including earrings and earphones, pose additional challenges. Unlike soft occluders such as hair, accessories introduce hard edges, specular highlights, and non-biological textures that can dominate learned features and bias similarity scores~\cite{emersic2017covariate}. To address these limitations, we present a fully automated accessory-aware diffusion inpainting framework with robust mask generation and cross-benchmark, multi-backbone evaluation.

\subsection{Inpainting and Generative Restoration for Biometrics}
Image inpainting has long been studied as a means of restoring missing or corrupted image regions. Classical approaches based on texture synthesis or exemplar matching can handle small gaps but struggle with large or structured occlusions~\cite{bertalmio2000image, criminisi2004region}. Deep learning–based inpainting methods, including encoder–decoder architectures and adversarial networks, have demonstrated improved realism and scalability~\cite{pathak2016context, yu2018generative, elharrouss2020image}, yet often produce over-smoothed results or hallucinated structures.

In biometric applications, restoration poses additional risks. Prior work in face biometrics has shown that generic inpainting may introduce subtle geometric distortions that alter identity-discriminative cues, leading to reduced recognition reliability~\cite{agarwal2024unmasking, ge2020occluded}. As a result, most biometric systems favor occlusion-robust feature extraction over explicit reconstruction, even though this may limit performance under severe occlusions.

For ear biometrics, only limited efforts have explored restoration-based approaches, typically relying on conventional inpainting or heuristic reconstruction methods~\cite{emervsivc2018towards}. These methods lack the capacity to recover fine anatomical details and do not explicitly optimize for downstream recognition performance.

\subsection{Diffusion Models for Image Inpainting}
Diffusion models have recently emerged as a powerful class of generative models, achieving state-of-the-art performance in image synthesis and editing~\cite{ho2020denoising, nichol2021improved}. By modeling the data distribution through an iterative denoising process, diffusion models are capable of producing high-fidelity images with strong global coherence and sharp local details. Conditional diffusion frameworks further enable guided generation for tasks such as image completion and inpainting~\cite{yu2023inpaint,saharia2022photorealistic, song2020score}.

Recent diffusion-based inpainting methods have demonstrated superior performance over GAN-based approaches, particularly for large missing regions and complex structures~\cite{saharia2022palette}. The iterative refinement process allows generated content to better align with surrounding context, reducing boundary artifacts and improving structural continuity. These properties are especially desirable for anatomically structured objects such as human ears~\cite{yu2023inpaint}.

\subsection{Positioning of This Work}
Our work connects diffusion-based image restoration with ear biometric recognition. Unlike prior occlusion-robust methods~\cite{emervsivc2018towards}, we explicitly reconstruct accessory-occluded ear regions using conditional diffusion inpainting. In contrast to generic inpainting approaches~\cite{saharia2022palette}, we incorporate structure-aware constraints tailored to ear anatomy, encouraging boundary continuity and texture consistency along key auricular components.

Importantly, we evaluate restoration not only in terms of visual quality but also by measuring its impact on downstream recognition performance across multiple datasets and transformer configurations. By formulating ear accessory removal as an anatomy-aware diffusion inpainting problem, our approach provides an effective preprocessing strategy for robust ear recognition in the presence of real-world accessory occlusions.


\begin{figure*}[t]
\centering
  \includegraphics[width=0.9\linewidth,clip=]{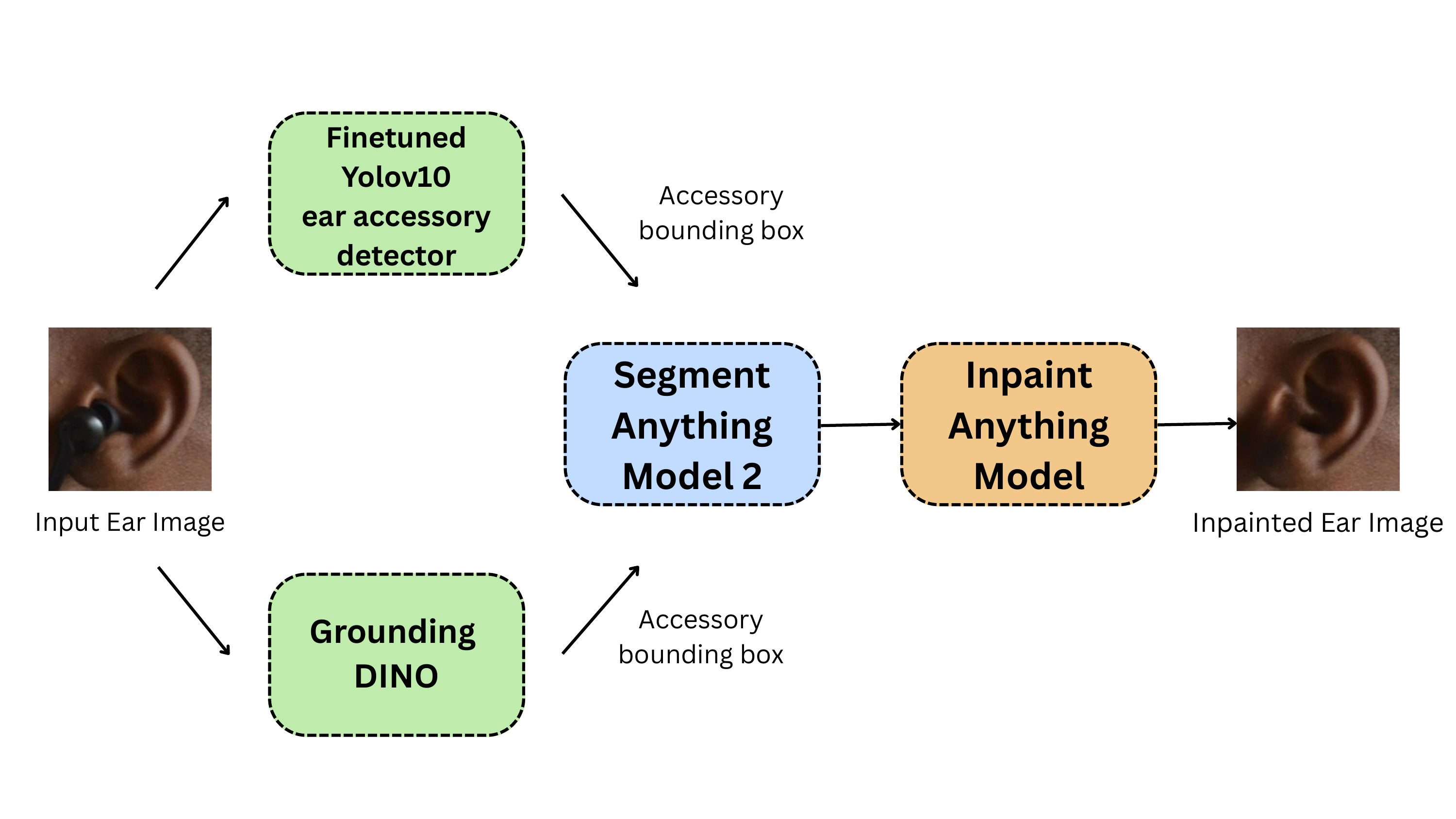}
  \caption{\textbf{Ear-accessory Inpainting pipeline.} Ear-accessory regions are localized in the input ear image using finetuned YOLOv10 accessory detector and Grounding DINO. The resulting bounding boxes are provided to Segment Anything Model 2 (SAM2) to produce accessory segmentation masks, which are post-processed and merged into a single binary mask. The mask along with the ear image is then sent to the Inpaint Anything model which gives the final inpainted ear image.}
  \label{fig:ear_accessory_pipeline}
\end{figure*}

\section{Methodology}
\label{sec:method}
\subsection{Overview}
Our pipeline removes ear accessories (e.g., earrings, earbuds, earphones) via two stages: (i) \textbf{accessory mask generation} and (ii) \textbf{masked restoration} (inpainting). Given an input ear image $\mathbf{I}$, we first localize potential accessories using a \emph{hybrid detector}- a supervised YOLOv10~\cite{wang2024yolov10} model and a zero-shot Grounding DINO detector~\cite{liu2023grounding}. The resulting bounding boxes are converted into high-quality pixel masks using SAM~2~\cite{ravi2024sam2}, followed by light morphological refinement. Finally, we inpaint the masked region to reconstruct an accessory-free ear image $\hat{\mathbf{I}}$ for downstream recognition. An overview of the complete pipeline is illustrated in Fig.~\ref{fig:ear_accessory_pipeline}.

\subsection{Accessory Localization}
\paragraph{Supervised detector (YOLOv10).}
We fine-tune YOLOv10~\cite{wang2024yolov10} on a manually annotated accessory dataset, where we label bounding boxes for common ear accessories (e.g., \emph{ear accessory}, \emph{earbud}) in ear images. At inference time, YOLOv10 produces candidate accessory boxes.

\paragraph{Zero-shot detector (Grounding DINO).}
To improve robustness to unseen accessory types and domain shift, we additionally use Grounding DINO~\cite{liu2023grounding} as a zero-shot detector. We query it with a curated text prompt comprising accessory phrases (e.g., `earring', `wireless earbud', `hearing aid'), formatted as lowercase, period-separated terms to match the model’s expected interface. The detector outputs candidate boxes along with confidence and text-alignment scores. We filter detections using the box and text thresholds, and discard overly large detections using an area-ratio constraint to avoid degenerate boxes spanning large portions of the ear region.

We obtain both YOLOv10 and Grounding DINO detections in our work: YOLOv10’s precision on known accessory categories while using Grounding DINO to capture rare or previously unseen accessories.

\subsection{Pixel-Accurate Mask Generation with SAM~2}
Bounding boxes provide only coarse localization, so we convert the proposals into pixel-level masks using SAM~2~\cite{ravi2024sam2}. For each detected box, we prompt SAM~2 with the box and enable multi-mask prediction. We binarize the masks, and merge them into a single accessory mask by taking the union across all accepted masks. If neither detector fires (or if the merged mask is empty), we save an empty mask to ensure that non-occluded images remain unchanged.

To reduce spurious speckles and stabilize mask boundaries, we apply light post-processing consistent with our code: a single erosion with a $3\times3$ kernel followed by median filtering (kernel size 5). This step improves mask compactness and suppresses small isolated regions.

\subsection{Masked Restoration via Diffusion-Based Inpainting}
Given an ear image and its predicted accessory mask, we reconstruct an accessory-free image using a diffusion-based inpainting model. We employ a LaMa-style inference pipeline, which synthesizes pixels inside the masked region while conditioning on the surrounding context~\cite{yu2023inpaint, suvorov2022resolution}. This encourages anatomically plausible completions and smooth continuity along key ear structures.

To support heterogeneous datasets, we enforce a consistent input format during inference. Specifically, we convert grayscale images to three-channel inputs by replication and drop alpha channels when present. If mask and image resolutions differ, we resize the mask using nearest-neighbor interpolation to preserve binary structure.

\subsection{Integration with Transformer-Based Ear Recognition}
We use the inpainted image as a drop-in pre-processing replacement for the original input in transformer-based ear recognition systems. All recognition settings (training, evaluation protocol, and similarity computation) are kept identical across experiments; only the pre-processing differs. We evaluate multiple vision transformer backbones and patch sizes to quantify how input tokenization granularity interacts with occlusion and how much benefit inpainting provides across architectures and datasets.

\subsection{Implementation Summary}
\begin{figure*}[t]
\centering
  \includegraphics[width=0.7\linewidth,clip=]{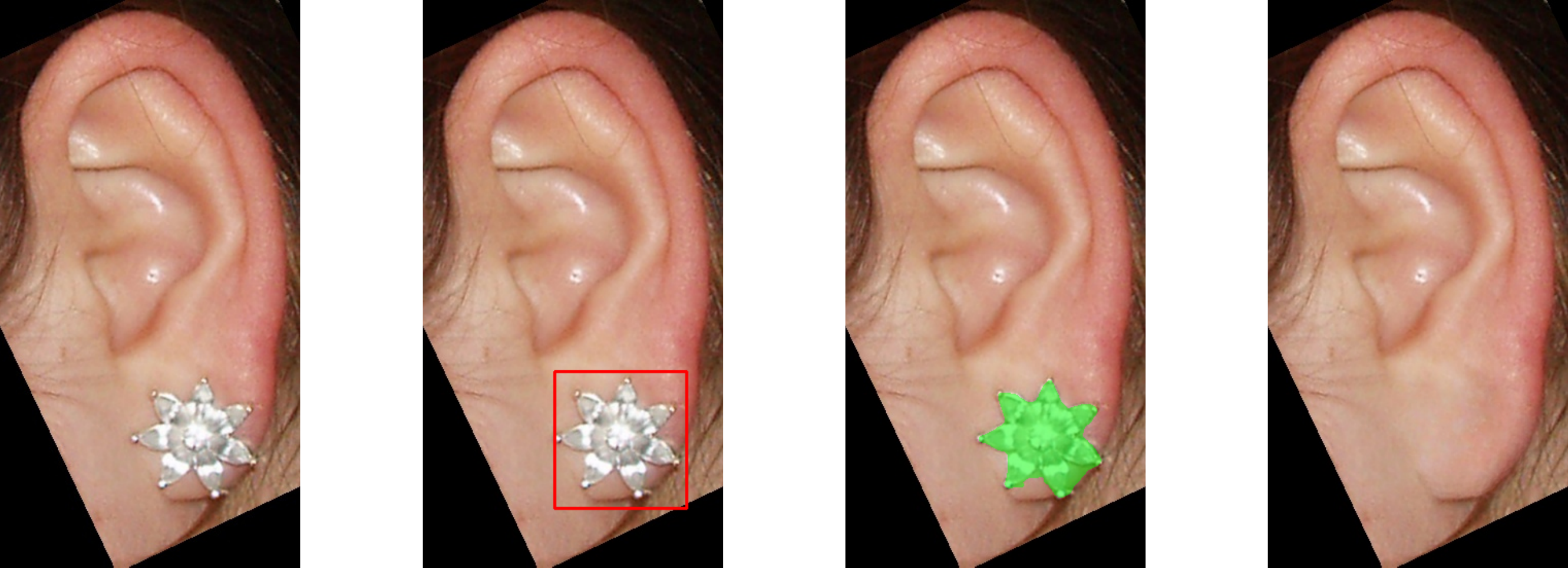}
  \caption{\textbf{End-to-End Ear Accessory Removal and Image Restoration.} From left to right: (1) input ear image, (2) detected accessory region highlighted with a red bounding box, (3) accessory segmentation mask in green, and (4) final ear image after accessory removal via inpainting. Accessory regions are detected using a fine-tuned YOLOv10 detector and Grounding DINO; the resulting boxes guide SAM2 to produce masks that are post-processed and merged into a single binary mask, which is then provided to Inpaint Anything to generate the inpainted ear image.}
  \label{fig:ear_accessory_images}
\end{figure*}

In practice, our pipeline proceeds as follows: (1) detect accessories with YOLOv10 and Grounding DINO, (2) obtain pixel masks with SAM~2 and refine them with light morphology, and (3) inpaint masked regions using the diffusion-based model with optional dilation. Fig.~\ref{fig:ear_accessory_images} illustrates an end-to-end example of these steps, from detection and mask generation to the final restored ear image. The resulting images are then forwarded to the recognition backbone.

\definecolor{impGreen}{RGB}{198,239,206}
\definecolor{closeOrange}{RGB}{255,229,204}

\renewcommand{\arraystretch}{1.5}
\begin{table*}[t]
\centering
\caption{\MakeUppercase{Comparison of AUC scores using non-overlapping ViT configurations.
Baseline results use original occluded ear images, while Inpainted results apply diffusion-based ear inpainting as a pre-processing step.
Cells in Orange indicate the closest score to the baseline within 3\% without surpassing it; cells in green indicate scores that surpass the baseline.}}
\label{tab:inpainting_results}
\begin{tabular}{lcccccc}
\toprule
\textbf{Model} & \textbf{Patch \& Stride} & \textbf{Input} &
\textbf{AWE} & \textbf{OPIB} & \textbf{WPUT} & \textbf{EarVN1.0} \\
\midrule

\multicolumn{7}{l}{\textit{Tiny (ViT\_T)}} \\
ViT\_T\_p16 & 16 & Baseline &
0.9832 $\pm$ 0.0013 & 0.9168 $\pm$ 0.0061 & 0.9442 $\pm$ 0.0055 & 0.7726 $\pm$ 0.0029 \\
ViT\_T\_p16 & 16 & Inpainted &
0.9260 $\pm$ 0.0020 &
\cellcolor{closeOrange}0.9098 $\pm$ 0.0018 &
0.9140 $\pm$ 0.0040 &
\cellcolor{impGreen}\textbf{0.7790 $\pm$ 0.0014} \\

ViT\_T\_p28 & 28 & Baseline &
0.9732 $\pm$ 0.0015 & 0.8926 $\pm$ 0.0055 & 0.9286 $\pm$ 0.0047 & 0.7356 $\pm$ 0.0017 \\
ViT\_T\_p28 & 28 & Inpainted &
0.9140 $\pm$ 0.0043 &
\cellcolor{closeOrange}0.8900 $\pm$ 0.0030 &
0.8784 $\pm$ 0.0059 &
\cellcolor{impGreen}\textbf{0.7510 $\pm$ 0.0020} \\

ViT\_T\_p56 & 56 & Baseline &
0.9250 $\pm$ 0.0019 & 0.8258 $\pm$ 0.0065 & 0.8782 $\pm$ 0.0082 & 0.6330 $\pm$ 0.0033 \\
ViT\_T\_p56 & 56 & Inpainted &
0.8790 $\pm$ 0.0060 &
\cellcolor{impGreen}\textbf{0.8510 $\pm$ 0.0050} &
0.8100 $\pm$ 0.0120 &
\cellcolor{impGreen}\textbf{0.6720 $\pm$ 0.0050} \\

\midrule
\multicolumn{7}{l}{\textit{Small (ViT\_S)}} \\
ViT\_S\_p16 & 16 & Baseline &
0.9778 $\pm$ 0.0013 & 0.9148 $\pm$ 0.0019 & 0.9382 $\pm$ 0.0024 & 0.7536 $\pm$ 0.0017 \\
ViT\_S\_p16 & 16 & Inpainted &
0.9200 $\pm$ 0.0050 &
\cellcolor{closeOrange}0.9040 $\pm$ 0.0030 &
0.9120 $\pm$ 0.0050 &
\cellcolor{impGreen}\textbf{0.7720 $\pm$ 0.0010} \\

ViT\_S\_p28 & 28 & Baseline &
0.9640 $\pm$ 0.0019 & 0.8934 $\pm$ 0.0076 & 0.9270 $\pm$ 0.0017 & 0.7164 $\pm$ 0.0017 \\
ViT\_S\_p28 & 28 & Inpainted &
0.9048 $\pm$ 0.0052 &
\cellcolor{closeOrange}0.8818 $\pm$ 0.0059 &
0.8840 $\pm$ 0.0040 &
\cellcolor{impGreen}\textbf{0.7396 $\pm$ 0.0005} \\

ViT\_S\_p56 & 56 & Baseline &
0.9132 $\pm$ 0.0041 & 0.8432 $\pm$ 0.0064 & 0.8864 $\pm$ 0.0040 & 0.6296 $\pm$ 0.0026 \\
ViT\_S\_p56 & 56 & Inpainted &
0.8734 $\pm$ 0.0059 &
\cellcolor{impGreen}\textbf{0.8520 $\pm$ 0.0040} &
0.8196 $\pm$ 0.0078 &
\cellcolor{impGreen}\textbf{0.6660 $\pm$ 0.0010} \\

\midrule
\multicolumn{7}{l}{\textit{Base (ViT\_B)}} \\
ViT\_B\_p16 & 16 & Baseline &
0.8828 $\pm$ 0.0194 & 0.8808 $\pm$ 0.0144 & 0.9244 $\pm$ 0.0111 & 0.7086 $\pm$ 0.0118 \\
ViT\_B\_p16 & 16 & Inpainted &
\cellcolor{impGreen}\textbf{0.9180 $\pm$ 0.0010} &
\cellcolor{impGreen}\textbf{0.9010 $\pm$ 0.0030} &
\cellcolor{closeOrange}0.9080 $\pm$ 0.0040 &
\cellcolor{impGreen}\textbf{0.7660 $\pm$ 0.0040} \\

ViT\_B\_p28 & 28 & Baseline &
0.9044 $\pm$ 0.0032 & 0.9036 $\pm$ 0.0040 & 0.9320 $\pm$ 0.0019 & 0.7278 $\pm$ 0.0033 \\
ViT\_B\_p28 & 28 & Inpainted &
\cellcolor{closeOrange}0.9010 $\pm$ 0.0060 &
0.8810 $\pm$ 0.0050 &
0.8780 $\pm$ 0.0070 &
\cellcolor{impGreen}\textbf{0.7294 $\pm$ 0.0030} \\

ViT\_B\_p56 & 56 & Baseline &
0.8796 $\pm$ 0.0072 & 0.8652 $\pm$ 0.0158 & 0.9144 $\pm$ 0.0053 & 0.6978 $\pm$ 0.0065 \\
ViT\_B\_p56 & 56 & Inpainted &
\cellcolor{closeOrange}0.8630 $\pm$ 0.0050 &
\cellcolor{closeOrange}0.8440 $\pm$ 0.0020 &
0.8230 $\pm$ 0.0040 &
0.6640 $\pm$ 0.0040 \\

\midrule
\multicolumn{7}{l}{\textit{Large (ViT\_L)}} \\
ViT\_L\_p16 & 16 & Baseline &
0.9712 $\pm$ 0.0018 & 0.9184 $\pm$ 0.0054 & 0.9364 $\pm$ 0.0043 & 0.7358 $\pm$ 0.0018 \\
ViT\_L\_p16 & 16 & Inpainted &
0.9200 $\pm$ 0.0041 &
\cellcolor{closeOrange}0.9040 $\pm$ 0.0040 &
0.9086 $\pm$ 0.0040 &
\cellcolor{impGreen}\textbf{0.7640 $\pm$ 0.0060} \\

ViT\_L\_p28 & 28 & Baseline &
0.9610 $\pm$ 0.0037 & 0.9028 $\pm$ 0.0042 & 0.9272 $\pm$ 0.0048 & 0.7176 $\pm$ 0.0046 \\
ViT\_L\_p28 & 28 & Inpainted &
0.8970 $\pm$ 0.0030 &
0.8770 $\pm$ 0.0020 &
0.8730 $\pm$ 0.0050 &
\cellcolor{impGreen}\textbf{0.7276 $\pm$ 0.0023} \\

ViT\_L\_p56 & 56 & Baseline &
0.9228 $\pm$ 0.0020 & 0.8574 $\pm$ 0.0092 & 0.9106 $\pm$ 0.0027 & 0.6606 $\pm$ 0.0030 \\
ViT\_L\_p56 & 56 & Inpainted &
0.8650 $\pm$ 0.0070 &
\cellcolor{closeOrange}0.8460 $\pm$ 0.0040 &
0.8310 $\pm$ 0.0080 &
\cellcolor{impGreen}\textbf{0.6678 $\pm$ 0.0015} \\

\bottomrule
\end{tabular}
\end{table*}

\section{DATASET}

\subsection{Training data}

The UERC2023 \cite{emervsic2023unconstrained} dataset was used to train the vision transformers. It is a composite data set assembled from the UERC2017 and UERC2019 competition data sets (14,004 images from 650 subjects) and the VGGFace-Ear~\cite{ramos2022vggface} dataset (234,651 images from 660 subjects).  The VGGFace-Ear dataset was created by cropping ear regions from face images in the VGGFace~\cite{cao2018vggface2} dataset and resizing them to a fixed size. We excluded images from the first 100 subjects from the UERC2023 dataset that also belonged to the AWE~\cite{emervsivc2017ear} dataset (used below for testing), so the final number of subjects was 2404, and the total number of images was 247,655. UERC2023 provides a wide variety of ear images encountered in real-world scenarios with variances in gender, ethnicity, poses, illumination and occlusion, making it suitable for model training.

\subsection{Testing data}

We used several distinct datasets for testing. The datasets used for testing include AWE, OPIB, WPUT, and EarVN1.0, with each dataset considered independently as a separate test set.

\begin{enumerate}

\item AWE (Annotated Web Ears)~\cite{emervsivc2017ear}  contains 1000 images -- exactly 10 images of each of 100 subjects. These images are close crops of the ears of famous persons and were collected through Web queries, which were then manually filtered. The gender breakdown is 91\% male and 9\% female. The ethnic composition is broad, with 61\% white, 18\% Asian, 11\% black, 3\% South Asian, 3\% Middle Eastern, 3\% South American, and 1\% other. For accessories, 91\% of the photographs show no accessories, 8\% show earrings, and 1\% show other accessories. Occlusion levels are classified as 65\% no occlusion, 28\% mild occlusion, and 7\% severe occlusion. The dataset also contains precise head position data with different distributions for pitch, roll and yaw. Furthermore, head side information is accessible, with 52\% labeled as Left and 48\% as Right. The dataset, therefore, includes a diverse collection of ear images with various attributes.

\item The OPIB dataset~\cite{Adebayo2023} contains 907 images from 152 persons of African ethnicity. Three images of the left ear and three of the right ear were captured from most subjects at 0, 30, and 60-degree angles relative to a profile view looking directly at the ear. The dataset contains ear images with occlusions, like headphones, scarves, and earrings. Both indoor and outdoor settings were used to gather the photos. Students from a Nigerian public university collected the data; the gender distribution was 59.65\% male and 40.35\% female. 

\item The West Pomeranian University of Technology (WPUT) dataset~\cite{frejlichowski2010west} contains 2071 cropped color images of the ears of 501 individuals. This dataset is rather diverse in environmental/imaging conditions, which were carefully captured and formed the basis for image file naming. Gender balance was approximately equal, with 50.70\% female and 49.30\% male. Images were taken indoors at 15.6\%, outdoors, and at 2\% dark. Two or more images of each of the left and right ears of each subject are provided in the dataset, with two choices of head pose (profile and 15 degrees frontal from profile). The distribution of ages was “0” to “above 50”, with the modal age being between 21 and 35 years of age. The dataset is dominated by subjects who appeared in only one imaging session. A significant proportion of the images (80\%) contain auricle deformations such as hair occlusions (166 subjects), earrings (147 subjects), spectacles, and other accessories, with the remaining 20\% devoid of occlusions. Motion blur artifacts were detected in 8\% of the photos, highlighting the difficulties in identifying the human ear under varying settings.

\item EarVN1.0~\cite{hoang2019earvn1} contains  28,412 unconstrained, low-resolution color ear images of 98 Asian men and 66 Asian women. The original facial images have been obtained in unconstrained conditions such as illumination, occlusion, and rotations, including camera systems and lighting conditions. Ear images are then cropped from facial images, featuring a variety of ear poses and illumination levels. Each subject is represented by 100 or more images in the dataset.

\end{enumerate}

\section{Experimental Setup}
\label{sec:exp_setup}

\subsection{Training Data and Pre-processing}
We train ear recognition models in a \emph{verification} setting using the UERC2023 training data, excluding the first 100 subjects that overlap with AWE. As UERC2023 contains both left- and right-ear images per subject, we first apply a ResNet-100-based ear-side classifier to separate left and right ears. Each ear side is treated as an independent identity, yielding two subject folders per original subject (left and right). We then align the ear images by segmenting the ear, estimating its vertical axis from the longest top-$k$ lines in the mask, rotating to make the axis upright, and cropping the segmented region. All input images are resized to $112\times112$ dimension.

\subsection{Backbones and Tokenization}
We evaluate four Vision Transformer backbones: ViT-T (Tiny), ViT-S (Small), ViT-B (Base), and ViT-L (Large). We use \emph{non-overlapping} patches (stride = patch size) and evaluate patch sizes ${16, 28, 56}$ to analyze how spatial tokenization affects occlusion robustness.
All model variants are trained from scratch for 100 epochs under the same protocol: AdamW (lr $1\times10^{-3}$, weight decay $0.1$), batch size $128$, FP16, and sample rate $0.3$, with a margin-based loss using margins $(1.0, 0.0, 0.4)$. Regardless of model scale, features are projected to a shared 512-D embedding space. We apply stochastic depth with variant-specific drop-path rates, and disable token masking during training.

\subsection{Evaluation Protocol}
We evaluate the trained models on four held-out benchmarks: AWE, OPIB, WPUT, and EarVN1.0. Following common biometric evaluation practice, we use each benchmark \emph{entirely for testing} (i.e., no train/test split within these datasets), since training is performed on UERC2023 and the goal is cross-dataset generalization.

We compare two input conditions. (i)
\textbf{Baseline (aligned and cropped):} models are trained on the aligned-and-cropped UERC2023 images and evaluated on the aligned-and-cropped versions of AWE, OPIB, WPUT, and EarVN1.0.
(ii) \textbf{Inpainted (alignedand cropped + inpainting):} we apply the proposed ear accessory inpainting module (Sec.~\ref{sec:method}) as a pre-processing step to the aligned-and-cropped images in \emph{both} training and testing scenarios. Models are trained on inpainted UERC2023 and evaluated on inpainted versions of AWE, OPIB, WPUT, and EarVN1.0.

For each ViT configuration and patch setting, baseline and inpainted models share the \emph{same architecture, training protocol, and hyperparameters}, enabling a direct attribution of performance differences to the pre-processing stage.

Ear matching was performed by computing the cosine
similarity scores of the 512-dimensional feature embeddings extracted from the ViT models. For a dataset with $N$ subjects and $M_i$ images per subject $i$, we form all genuine and impostor pairs:
\begin{equation}
\#\text{genuine}=\sum_{i=1}^{N} \binom{M_i}{2},\qquad
\#\text{impostor}=\sum_{i=1}^{N-1}\sum_{j=i+1}^{N} M_iM_j .
\end{equation}
Performance is summarized using the Area Under the ROC Curve (AUC), which reflects the separability between genuine and impostor score distributions. Each matching experiment is repeated five times to account for stochasticity (e.g., initialization and training dynamics). We report the mean AUC with empirical variability estimates in Table~\ref{tab:inpainting_results}.

\section{Results}
\label{sec:results}

This section evaluates diffusion-based ear inpainting as a pre-processing step for transformer-based ear recognition. We report AUC across four datasets (AWE, OPIB, WPUT, EarVN1.0) and analyze performance as a function of ViT scale (Tiny/Small/Base/Large) and non-overlapping patch size (16/28/56). 

\subsection{Impact of Patch Size}
\label{subsec:patch_size}

The effect of inpainting is strongly coupled with the ViT patch size. At the coarsest setting (patch size 56), inpainting yields the most reliable gains, particularly on OPIB and EarVN1.0 across model scales. This behavior aligns with the tokenization mechanism: under coarse sampling, a single occluder can corrupt entire patches, removing most discriminative evidence. Inpainting partially restores meaningful ear structure within those patches, producing more informative token representations for the transformer.

At finer patch sizes (16 and 28), baseline models already benefit from denser spatial sampling, which reduces the impact of localized occlusions. In these settings, inpainting is more mixed: it can improve robustness on challenging imagery (notably EarVN1.0), but may also slightly reduce performance on cleaner datasets, suggesting that generative reconstruction can occasionally perturb subtle identity cues. Overall, the gains are most pronounced when coarse tokenization amplifies occlusion damage, whereas fine tokenization provides inherent resilience.

\subsection{Dataset-Specific Performance}
\label{subsec:dataset_analysis}

\textbf{EarVN1.0} benefits most consistently from inpainting across nearly all model scales and patch sizes. As this dataset includes unconstrained imagery with significant occlusions, pose variation, and background clutter, restoration tends to remove non-biometric artifacts and recover missing ear structure, leading to repeated AUC improvements across configurations.

\textbf{OPIB} shows moderate but meaningful sensitivity to inpainting. While gains are not uniform across all settings, improvements are most evident for larger patches, where inpainting counteracts the degradation observed in the baseline under coarse spatial representations.

\textbf{AWE and WPUT} contain more controlled or semi-controlled images with comparatively fewer severe occlusions. Baselines therefore attain high AUC at fine patch sizes, leaving limited headroom for restoration. In these regimes, inpainting occasionally provides marginal gains but can also yield small drops, indicating that when the occlusion burden is low, reconstructing may introduce unnecessary variation.

\subsection{Effect of Model Scale}
\label{subsec:model_scale}

Table~\ref{tab:inpainting_results} reports AUC scores for baseline occluded inputs and diffusion-inpainted pre-processing across non-overlapping ViT configurations and datasets. On EarVN1.0, inpainting improves 11/12 configurations with a mean gain of +0.0175 AUC over the baseline mean (0.7074), i.e., \(\sim\)2.5\% relative. The largest improvement occurs for ViT\_B\_p16 (0.7086 \(\rightarrow\) 0.7660; +0.0574 AUC, +8.1\%), with strong gains also under coarse tokenization (ViT\_T\_p56: +0.0390, +6.2\%; ViT\_S\_p56: +0.0364, +5.8\%). Overall, the most favorable regime is ViT\_B\_p16, which improves AWE (+4.0\%), OPIB (+2.3\%), and EarVN1.0 (+8.1\%) while incurring only a modest WPUT decrease (-1.8\%), indicating that sufficient capacity with fine patches can best exploit inpainted content while limiting its downsides.

\subsection{Trade-offs and Failure Modes}
\label{subsec:failure_modes}

While diffusion restoration improves robustness under severe occlusions, it can underperform on cleaner inputs. A plausible explanation is that diffusion priors favor anatomically plausible synthesis, which may smooth or alter fine-grained identity cues (e.g., local folds and ridge patterns) that are important for matching. This effect is most visible when baseline performance is already near saturation (e.g., AWE/WPUT with small patches), where small reconstruction deviations can outweigh the benefits of artifact removal. 

\begin{figure*}[t]
\centering
  \includegraphics[width=0.65\linewidth,clip=]{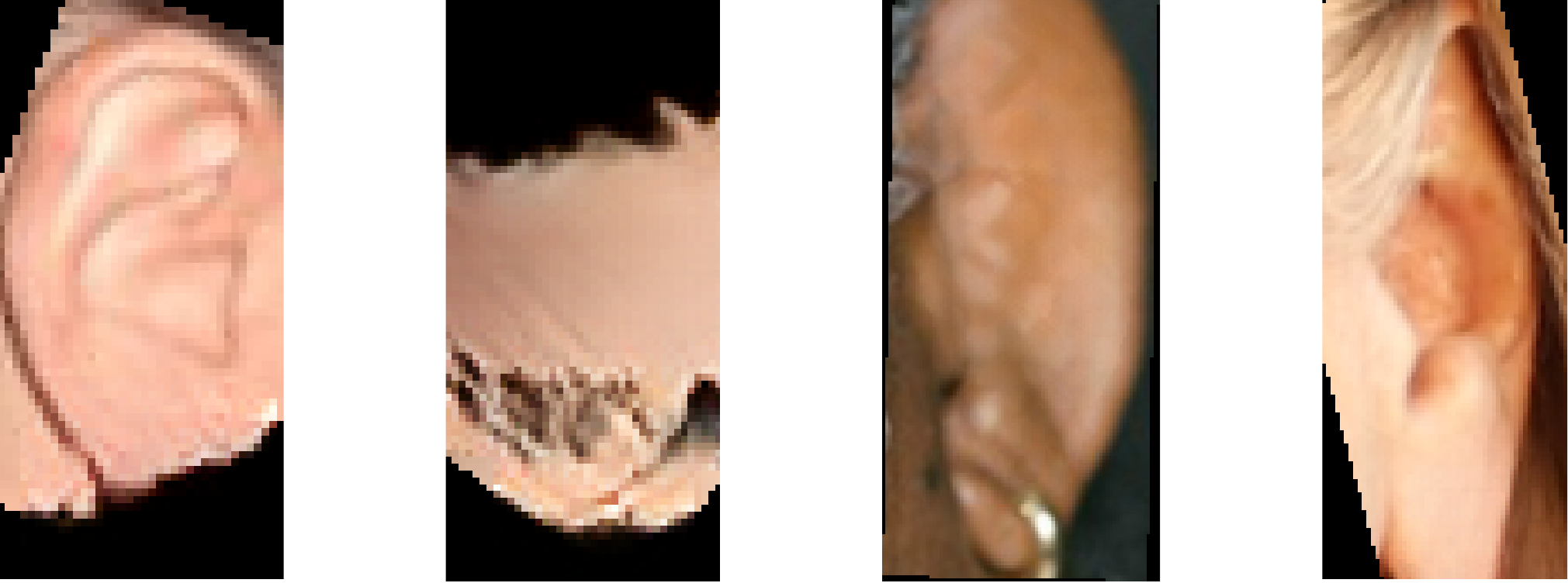}
  \caption{\textbf{Failure cases of inpainting.} Inpainting can fail when the accessory mask is inaccurate or when black regions introduced by rotating ear images bias restoration, leading to artifacts, distorted reconstructions, and occasionally stretched ear lobules.}
  \label{fig:ear_accessory_failure}
\end{figure*}

Inpainting quality can degrade when the accessory mask is inaccurate or when image rotation introduces black padded regions that bias restoration, occasionally producing visible artifacts, distorted geometry, or stretched lobule structures, which leads to a drop in the recognition performance. Figure~\ref{fig:ear_accessory_failure} further illustrates representative failure cases. 

These observations motivate identity-preserving restoration as a key direction: incorporating feature-level consistency constraints with a frozen recognizer, occlusion-aware guidance, or confidence-based gating (apply inpainting only when the confidence of occlusion is detected is higher) could reduce unnecessary reconstruction and improve stability across datasets.

\subsection{Discussion and Practical Implications}
\label{subsec:discussion}

From a deployment perspective, diffusion-based inpainting is most attractive when (i) occlusions are frequent or high-contrast (earphones, earrings), (ii) resource constraints favor coarse patching or lightweight backbones, or (iii) the target domain resembles unconstrained imagery. In these conditions, inpainting acts as an information recovery module that improves the effective input quality without changing the recognition architecture.

Conversely, when imagery is clean and baselines already saturate, unconditional inpainting may be unnecessary and can slightly reduce accuracy. A practical strategy is to combine restoration with occlusion detection or uncertainty estimation to selectively invoke inpainting only when it is likely to help.

\subsection{Summary of Results}
\label{subsec:summary}

In summary, diffusion-based inpainting:
\begin{itemize}
    \item Provides the most reliable improvements under coarse patching (patch size 56), where occlusions can corrupt entire tokens.
    \item Delivers the most consistent gains on unconstrained data with strong occlusions (EarVN1.0), indicating robustness to real-world conditions.
    \item Complements transformer-based recognition by restoring semantically meaningful patch content prior to attention-based aggregation.
    \item On cleaner datasets, using smaller patches gives mixed results, showing a trade-off: removing artifacts better can also harm identity preservation.

\end{itemize}

Overall, the results support diffusion-based ear restoration as a practical robustness mechanism for occlusion-heavy scenarios, while also emphasizing the importance of identity-preserving and selective inpainting for broader reliability.

\section{CONCLUSIONS AND FUTURE WORKS}

We introduced a diffusion-based ear inpainting framework that explicitly addresses a practical and underexplored source of failure in ear biometrics: occlusions caused by accessories such as earrings and earphones. The proposed method addresses this by considering the removal/reconstruction of accessories as a generative reconstruction problem by utilizing strong priors of diffusion model for anatomical region reconstruction while being consistent with the surrounding visible context. This preprocessing step is model-agnostic and can be readily paired with standard recognition backbones, making it a practical solution for unconstrained deployments where accessory-induced artifacts are common.

Comprehensive experiments across multiple benchmarks and non-overlapping ViT configurations show that diffusion-based restoration is most beneficial under challenging conditions, where occlusions are large, high-contrast, or cover discriminative ear components. In these regimes, inpainting produces consistent AUC gains -- most notably on EarVN1.0 across all evaluated model scales and patch sizes--indicating improved robustness when the input is heavily corrupted. At the same time, we observe that on less challenging datasets and finer patch settings, generative reconstruction can occasionally reduce performance, suggesting that even visually plausible synthesis may alter subtle identity cues important for recognition. These results highlight an important takeaway: diffusion priors are effective at removing non-biometric artifacts, but identity preservation must be treated as a first-class objective when restoration is used for biometric matching.

This work opens several promising directions. First, incorporating explicit identity-preserving constraints--through feature-level consistency with a frozen recognizer, perceptual/metric losses, or conditional guidance--can help prevent over-smoothing and preserve fine-grained morphology. Second, extending the restoration model to handle mixed and compound occlusions (e.g., hair, scarves, shadows, and sensor noise) would broaden applicability in the wild since existing diffusion-based inpainting methods are largely general-purpose and prioritize perceptual realism rather than task-specific objectives. When applied directly to biometric preprocessing, they do not guarantee preservation of identity-relevant geometry, particularly in regions critical for recognition. Finally, tighter integration of restoration and recognition, either through joint training or feedback-driven optimization, may enable the inpainting module to adapt its reconstructions to what the matcher needs rather than what looks most realistic. Overall, our study demonstrates that diffusion-based inpainting is a viable and scalable pathway to improve ear recognition robustness in the presence of accessories, and it provides clear evidence and insights for building the next generation of occlusion-aware biometric pipelines.

{\small
\bibliographystyle{ieee}
\bibliography{egbib}
}

\end{document}